\documentclass[conference]{IEEEtran}
\IEEEoverridecommandlockouts

\usepackage[T1]{fontenc}
\usepackage{graphicx}
\usepackage{amsmath}
\usepackage{amssymb}
\usepackage{booktabs}
\usepackage{enumitem}
\usepackage{caption}
\usepackage{url}
\usepackage{cite}
\usepackage{xcolor}
\usepackage{hyperref}

\hypersetup{
  hypertexnames=false,
  colorlinks=true,
  linkcolor=blue!50!black,
  citecolor=blue!50!black,
  urlcolor=blue!50!black
}

\usepackage{fancyhdr}
\newcommand{\footerlabel}{Preprint, arXiv v2 (November 2026)}
\fancypagestyle{plain}{\fancyhf{}\fancyfoot[C]{\footnotesize \footerlabel \hfill \thepage}}
\newtheorem{definition}{Definition}

\begin{document}

\title{From Structured Graphs to Unstructured Evidence:\\
Agentic Root Cause Analysis on Digital Twins}

\author{\IEEEauthorblockN{Nicolas Tacheny}
\IEEEauthorblockA{University of Mons (UMONS), Belgium \\
Ni2 Innovation Lab, Li\`ege, Belgium \\
nicolas.tacheny@\{umons.ac.be,ni2.com\}}}

\maketitle
\thispagestyle{plain}  

\begin{abstract}
Root cause analysis over telecom and datacenter infrastructures is commonly addressed by methods that consume structured data, from graph traversal and rule-based correlation to probabilistic and learned models. Yet a digital twin of such an infrastructure cannot model everything at a reasonable cost. Operators therefore record much of what changes as unstructured data attached to the graph, where the discriminating evidence often sits. Methods restricted to structured attributes cannot use it. A diagnosis needs a layer that interprets this data. An agentic layer is a natural candidate, since which fragment matters depends on hypotheses that emerge during the investigation. We demonstrate its feasibility with ReAct, the most widely used agentic architecture. A single LLM interprets the descriptions and explores the graph through a typed tool interface exposed via the Model Context Protocol (MCP). On a benchmark of nine synthetic scenarios covering single-node, multi-node and distractor-heavy evidence (360 runs across four LLMs), a frontier-tier model names the root cause and its type on every run. It reaches 96.7\% end-to-end diagnostic accuracy, which also requires the exact supporting evidence. A mid-tier model reaches 86.7\%. Accuracy drops for every model when the evidence must be combined across several resources.
\end{abstract}


\section{Introduction}

Telecom and datacenter operators rely on highly interconnected infrastructures of physical and logical resources that deliver services to customers. When a service degrades, engineers must identify the resource responsible, a task known as root cause analysis (RCA). The standard tool kit reads structured data. It ranges from dependency-graph traversal and rule-based correlation to probabilistic attribution and graph neural networks~\cite{dang2019aiops,notaro2021aiops,chen2023balance,wang2023reason}. Mature as these approaches are, they assume that the information discriminating between candidate causes is encoded in the graph's structured attributes, such as node and edge types, status fields or identifiers.

In operational settings this assumption fails. A digital twin of the infrastructure structures what recurs and leaves the rest unstructured, because every class added to its data model is one more that its users must understand and keep current. Much of what lets an engineer identify a root cause therefore sits in unstructured data attached to the graph, most often natural-language fragments. An event description may read ``upgraded to a beta software version on Tuesday'', a note may record ``switch replaced last week''. Two events on the same resource may be structurally indistinguishable while their descriptions tell different stories, one routine and the other causal.

When the decisive information is absent from the structured attributes, classical algorithms offer no guarantee of finding the cause among the candidates reachable from the affected service. Ranking those candidates requires interpreting the unstructured data. An agentic layer is a natural way to do it, because which fragment matters depends on hypotheses that emerge during the investigation. Such a layer can choose what to read next as those hypotheses form. We take the most widely used agentic architecture, ReAct~\cite{yao2022react}, as an instance. In it a single LLM alternates between reasoning traces and tool calls over a typed tool interface exposed via the Model Context Protocol (MCP)~\cite{mcp-spec}. ReAct serves here as an instance of the agentic layer, not as a candidate for the best one.

We make three contributions. First, we characterise a regime of RCA that structure-only methods cannot address, because the discriminating evidence lives in unstructured data attached to the graph. Second, we show that an agentic layer is a feasible way to interpret that data, with ReAct as an instance served by a typed MCP tool interface that makes every step of the agent checkable. Third, we contribute a benchmark of nine synthetic scenarios in three categories, defined by where the discriminating evidence sits, with strict and machine-checkable scoring.


\section{Related Work}
\label{sec:related_work}

Three neighbouring literatures frame the paper, on telecom information models, on RCA in operational settings and on algorithmic and hybrid RCA. The tool-augmented agent paradigm we leverage is introduced where it is used (Section~\ref{sec:agentic-instantiation}).

Our abstraction into \emph{services}, \emph{resources} and \emph{events} follows the layering of the TM Forum Information Framework (SID) \cite{itu-m3190,tmforum-sid}, which underpins modern OSS/BSS systems. SID-style models say little about the natural-language descriptions attached to events and notes, which is the very content this paper focuses on. We inherit the shape of our tool-level primitives from SID and treat the unstructured fields it leaves underspecified as first-class evidence.

RCA in telecom and microservice infrastructures is a longstanding problem made non-trivial by alarm storms and configuration drift. Automated correlation is widely seen as a key enabler for service assurance \cite{zhang2021influence}. It sits within the broader AIOps research agenda, structuring failure detection, prediction, RCA and remediation as an integrated lifecycle \cite{dang2019aiops,notaro2021aiops}. Recent surveys call for formal frameworks beyond heuristic symptom matching \cite{fang2025goal,wang2024comprehensive}. Across this literature, the unstructured data operators routinely produce, technician notes and post-mortem reports among it, is typically treated as preprocessing input, outside the diagnostic loop. LLMs have recently entered that loop for cloud incidents. RCACopilot~\cite{chen2024rcacopilot} gathers diagnostic data through handlers predefined per alert type, then asks an LLM for the root cause category and an explanation. RCAgent~\cite{wang2024rcagent} lets a tool-augmented LLM agent collect and analyse incident data on its own. Both target cloud incidents. We target the unstructured data attached to the nodes of a typed infrastructure graph and ask what methods confined to its structure cannot do there.

Beyond traversal and rule-based correlation, recent work explores Bayesian attribution \cite{chen2023balance} and hierarchical GNNs over learned causal graphs \cite{wang2023reason}, often reported as one of the strongest baselines on structured-input RCA, with empirical studies reporting limited cross-system generalisation \cite{pham2024causalrca}. These approaches consume structured inputs only, such as KPI series or typed edges. Their performance is therefore bounded by the expressiveness of those features. A natural objection is that unstructured operational data could first be externalised into structured features by an NLP pipeline and handed back to such an algorithm. Hybrid neuro-symbolic systems take this route in adjacent settings \cite{garcez2023neurosymbolic}, as does recent work on knowledge graphs for autonomous driving \cite{wickramarachchi2024knowledgegraphsdrivingscenes}. Such a pipeline has to anticipate which fragments are relevant, which is precisely what the diagnostic process determines.


\section{Problem Setting}
\label{sec:problem-setting}

A telecom or datacenter infrastructure can be pictured as two stacked layers. On top is a hierarchy of \emph{services} delivered to customers, such as a mobile data plan or a transaction processing service. Below is a pool of \emph{resources} that implement them, such as virtual machines or databases. A single resource may serve several services at once, while a single service typically rests on many resources. Operations teams continuously annotate this graph with operational records, from technician notes after maintenance to change tickets and post-incident reports. We call the structured-plus-annotated representation that results a \emph{digital twin}~\cite{tao2018digital}, an ontological model of the live system that operations teams query and update as a proxy for the system itself. A full digital twin also follows the live system through its lifecycle and may simulate it. None of this is exercised here. This paper uses the twin's data model alone, a typed directed graph described by an ontology.

That the twin keeps part of its content unstructured is a modelling choice made under pressure. Every class added to its data model is one more that its users must understand and keep current. A change of version of an application installed on a server shows the cost. Modelling it takes classes for installed software and its versions, relations to the hosts and a procedure that operators follow at each upgrade, while a technician records the same fact in one sentence. A workable twin thus structures what recurs and leaves the rest unstructured, which is often where the deciding evidence of a diagnosis sits.

When a customer-facing service degrades, the engineer looks for the resource or set of resources responsible among those that implement it, comparing each candidate against the natural-language information attached to it. A one-line note ``upgraded to a beta software version on Tuesday'' is enough to tell apart two resources with otherwise identical structured attributes. So is an event reporting an unscheduled power cycle. RCA on a digital twin is therefore a graph traversal in which the deciding information at each visited node is unstructured.

We study a deliberately narrow class of instances. The ontology is minimal and fixed, an incident on one named service triggers each query, the cause-type taxonomy is closed and the discriminating evidence sits in the natural-language fields of nodes within a few hops of the incident. Evolving infrastructures, stale or contradictory evidence and multi-incident correlation are out of scope.


\section{Information Model}
\label{sec:information-model}

We model the digital twin as a typed directed graph, the \emph{infrastructure ontology}, restricted to the subset of the TM Forum SID layering~\cite{itu-m3190,tmforum-sid} that the regime requires. The word \emph{ontology} is used in the sense of OSS/BSS information models, a typed schema of classes and relations, not a description-logic artefact. Its four classes and four relations form the schema, of which each benchmark scenario is one instance. Nothing in the method depends on an ontology language. Any relational schema with the same types and the same tools would serve equally.

A \emph{service} is delivered to customers and may decompose into sub-services. Each \emph{resource}, a physical or logical component such as a virtual machine or a database, implements one or more services. An \emph{event} is a time-bounded occurrence on a resource, such as a maintenance window or an incident. A \emph{note} is an annotation on a resource, with the same natural-language $\textit{description}$ field as an event. Upward edges link a resource to the service it $\textsc{Implements}$ and a sub-service to the parent it is a $\textsc{ServiceOf}$. In the other direction, a resource is $\textsc{AffectedBy}$ an event and $\textsc{HasNote}$ a note, both edges pointing down to the fragment it carries.

\begin{definition}[Typed infrastructure ontology]
\label{def:ontology}
The infrastructure is modeled as a typed directed graph \[ G \;=\; (V,\, E,\, \tau_V,\, \tau_E) \] where each node $v \in V$ has a type \[ \tau_V(v) \;\in\; \mathcal{N} \;=\; \{\textsc{Service}, \textsc{Resource}, \textsc{Event}, \textsc{Note}\} \] and each edge $e = (u, v) \in E$ has a semantic type \[ \begin{aligned} \tau_E(e) \;\in\; \mathcal{E} \;=\; \{ &\textsc{Implements}, \textsc{AffectedBy}, \\ &\textsc{ServiceOf}, \textsc{HasNote}\}. \end{aligned} \] Each node carries an identifier, a name and its type.
\end{definition}

\noindent
The candidates of an RCA query on service $s$ are the resources of its \emph{implementation} $\sigma(s)$, those from which a directed path of $\textsc{Implements}$ and $\textsc{ServiceOf}$ edges leads up to $s$.


\section{Method}
\label{sec:method}

This section shows that a class of RCA instances is out of reach for any algorithm restricted to $(V, E, \tau_V, \tau_E)$, then describes the agentic instantiation we evaluate.

\subsection{Where Classical Graph Algorithms Break Down}
\label{sec:decomposition}

We describe a diagnosis as a walk over $G$ that starts from the service named in the incident and ends when a diagnosis is committed. The walker maintains a \emph{hypothesis state}, an evolving set of candidate causes and the fragments supporting them, left informal here. At each visited node an \emph{interpretation step} reads the fragments attached to it and updates the state. An \emph{exploration step} then picks the next node from the updated state. The choice of the next node is thus mediated by an interpretation of unstructured data, against everything accumulated so far.

Two regimes of RCA require this mediation. \emph{(i) Local interpretation gates exploration.} A resource may carry three maintenance events with similar structured fields, two reading ``routine maintenance, no impact'' and one reading ``upgraded to a beta software version on Tuesday morning''. Any topological policy that ignores descriptions cannot distinguish them and has no reason to next visit other resources subject to the same upgrade. An interpretation step does. \emph{(ii) Some causes are only legible across multiple nodes.} A note ``switch replaced last week'' on one resource and an event ``intermittent packet loss since Tuesday'' on a downstream resource are each individually unremarkable. Together they identify the cause. Recovering this requires interpretation to operate on the \emph{accumulated} state, not just the local fragments at the current node.

Both regimes defeat classical RCA, whether graph traversal, rule-based correlation or GNNs over structured attributes, since all of them assume the discriminating information is encoded in $(V, E, \tau_V, \tau_E)$. Regime (i) makes successor choice depend on content absent from any structured edge. No purely local summary survives regime (ii). An upstream stage could externalise the fragments into structured features or collect them all for a single model call. Both are interpretation layers in their own right. They fix in advance what to extract or collect, where an agent decides as its hypotheses form. Our benchmark does not separate these options, as Section~\ref{sec:discussion} discusses.

\smallskip
\noindent\textbf{Observation (informal).} \emph{Two RCA instances can share the same graph, the same structured attributes and the same incident query while their descriptions point to different root causes. Any algorithm operating solely on $(V, E, \tau_V, \tau_E)$ gives the same answer on both and is wrong on at least one. No such algorithm is correct on every instance of the regime.}

\subsection{Agentic Instantiation}
\label{sec:agentic-instantiation}

Both steps run within a single ReAct agent~\cite{yao2022react}. At each turn the model emits a reasoning trace and a tool invocation, whose output is appended to the conversation, until a terminal $\textsc{Publish}$ call ends the loop. The graph is exposed through five typed tools served via MCP~\cite{mcp-spec}, the only channel through which the agent obtains data from $G$. An identifier the agent mentions without a prior call returning it is therefore a hallucination, which the MCP record makes detectable.

\begin{definition}[MCP tool interface]
\label{def:mcp-interface}
The MCP server exposes five typed tools over $G$.
\begin{itemize}
 \item $\textsc{LookupService}(name) \to s$ returns the service matching a name or identifier.
 \item $\textsc{GetImplementation}(s) \to \sigma(s)$ returns the resources implementing $s$.
 \item $\textsc{GetNotes}(r)$ and $\textsc{GetEvents}(r)$ return the notes and events of resource $r$, each with its identifier and description.
 \item $\textsc{Publish}(\mathit{rootCauseIds},$\linebreak[1] $\mathit{rootCauseType},$ $\mathit{evidenceIds},$ $\mathit{summary})$ ends the investigation with the root cause resources, a cause type from the closed taxonomy \{\texttt{software\_update}, \texttt{configuration\_update}, \texttt{hardware\_anomaly}, \texttt{capacity\_change}, \texttt{unknown}\}, the supporting notes and events, and a summary.
\end{itemize}
\end{definition}

\noindent
The graph is a JSON document loaded by the MCP server for each run. Dedicated tools keep every call within one relation of Definition~\ref{def:ontology}, which a general query language such as SPARQL would not. Each call can then be checked against the MCP record.

The system prompt of the agent, not a workflow, prescribes a five-step protocol. It has the agent extract the service name from the incident, resolve it and retrieve $\sigma(s)$, then read the notes and events of each resource before calling $\textsc{Publish}$. Its reading of the descriptions is the interpretation step, its choice of the next tool call the exploration step. Because the protocol visits every resource of $\sigma(s)$, our benchmark exercises interpretation and synthesis more than the choice of what to read.


\section{Benchmark}
\label{sec:benchmark}

The benchmark has nine synthetic scenarios in three categories that follow the regimes of Section~\ref{sec:decomposition}, with the discriminating evidence at a single node (A), spread across several nodes (B) or hidden among distractors (C). It is built to demonstrate the claim, not to estimate performance on production incidents. Each scenario is a tuple $(G, q, R^*, E^*, T^*)$ of an ontology $G$, an incident message $q$ and three pieces of ground truth, namely the root cause set $R^* \subseteq V$, the causal fragments $E^*$ and the cause type $T^*$ from the taxonomy of Definition~\ref{def:mcp-interface}. Within the implementation of the queried service the candidates carry the same kinds of attributes. Only the descriptions say which one failed.

\begin{figure}[t]
\centering
\includegraphics[width=\columnwidth]{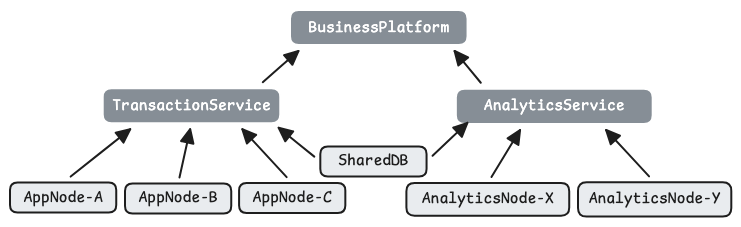}
\caption{Common ontology $G_0$ shared across all nine benchmark scenarios.}
\label{fig:ontology}
\end{figure}

All scenarios share the ontology $G_0$ of Figure~\ref{fig:ontology}. A root \texttt{BusinessPlatform} decomposes into \texttt{TransactionService} and \texttt{AnalyticsService}, with three and two dedicated resources respectively, plus a sixth resource \texttt{SharedDB} implementing both. Only the incident query and the text attached to the six resources vary. One structural shortcut remains. As the only resource shared by both sub-services, \texttt{SharedDB} is in $R^*$ on every scenario, which lets the query target either side without changing the cause. A rule that always names it would find the faulty resource on eight scenarios out of nine, though it would have to guess the evidence and the cause type. The benchmark therefore measures whether an agent reads the data correctly. It does not measure how far reading beats such a prior. In B2 $R^*$ also holds a co-located resource, because the fault sits in a rack the two share.

Within each category the three scenarios vary the form of the evidence. Category A places one causal fragment among routine siblings (A1), as an unsanctioned change that does not read as a fault (A2) or in operator language with no obvious keyword (A3). In category B the cause takes a temporal correlation across two nodes (B1), a shared rack known only from two notes (B2) or a migration legible only from three notes (B3). Distractors in category C bury the cause among trivial notes (C1), among alarming but irrelevant ones (C2) or next to a distractor on the same resource (C3). One scenario per category is given verbatim below, causal fragments in \textbf{bold}, with \texttt{[N]} marking a note and \texttt{[E]} an event. All nine will be released with the code upon publication.

\begingroup
\footnotesize
\linespread{0.95}\selectfont
\begin{description}[leftmargin=1.5em, labelindent=0pt, labelsep=0.3em, itemsep=1pt, parsep=0pt, topsep=1pt, font=\normalfont\bfseries]
\item[A1.] \emph{Software-beta upgrade.} Causal event among siblings of the same form on \texttt{SharedDB}. [E] \texttt{SharedDB}: ``Routine vacuum and statistics refresh completed without incident.'' [E] \texttt{SharedDB}: \textbf{``Database engine upgraded to a beta software version 4.2.1-rc on Tuesday morning, post-upgrade verification skipped.''} [E] \texttt{SharedDB}: ``Routine log rotation completed without incident.'' \\ \emph{Query:} ``\texttt{TransactionService} is degraded since Tuesday.'' \\ \emph{Cause type:} \texttt{software\_update}.
\item[B2.] \emph{Shared physical context.} Rack and PDU are not in $G_0$ and exist only as content of two notes. $R^* = \{\texttt{SharedDB}, \texttt{AppNode-A}\}$. [N] \texttt{AppNode-A}: \textbf{``Primary transactional node, hosted in rack RU-7, last hardware audit clean.''} [N] \texttt{AppNode-B}: ``Hosted in rack RU-12, healthy.'' [N] \texttt{AppNode-C}: ``Hosted in rack RU-15, healthy.'' [N] \texttt{SharedDB}: \textbf{``Shared SAN-backed database, physically located in rack RU-7. Recent power audit flagged a faulty PDU on this rack, replacement scheduled but not yet completed.''} \\ \emph{Query:} ``\texttt{TransactionService} experiences sporadic write failures on \texttt{AppNode-A}.'' \\ \emph{Cause type:} \texttt{hardware\_anomaly}.
\item[C2.] \emph{Alarming-but-irrelevant distractors.} The distractors are substantive themselves and the test is to match fragments with symptom timing. [N] \texttt{AppNode-A}: ``Suffered a memory leak two months ago, fully patched, monitored for recurrence; no further incidents.'' [N] \texttt{AppNode-C}: ``Was decommissioned and replaced last year during the datacenter migration, current hardware is the replacement unit.'' [N] \texttt{AnalyticsNode-X}: ``Recently flagged in capacity-planning report as approaching CPU saturation under projected load growth.'' [N] \texttt{AnalyticsNode-Y}: ``Standard analytics node, no notable history.'' [E] \texttt{AppNode-B}: ``Logged an unauthorised SSH attempt last quarter, source IP blocklisted, audit closed.'' [E] \texttt{SharedDB}: \textbf{``Storage volume expanded from 4TB to 8TB on Friday afternoon, automatic re-indexing job started but its progress is not currently reported by the database engine.''} \\ \emph{Query:} ``\texttt{TransactionService} queries have become significantly slower since Friday evening.'' \\ \emph{Cause type:} \texttt{capacity\_change}.
\end{description}
\endgroup


\section{Experiments}
\label{sec:experiments}

\subsection{Models}
\label{sec:models}

We evaluate four LLMs with native tool-calling. \textbf{Claude Opus 4.7} and \textbf{Claude Haiku 4.5} (Anthropic) are a frontier-tier and a compact mid-tier proprietary model. Two open-weight models of different scales, \textbf{Llama 3.3 70B Versatile} and \textbf{GPT-OSS-120B}, are both served by Groq. Explicit reasoning configurations such as extended thinking are out of scope.

\subsection{Evaluation Protocol}
\label{sec:eval-protocol}

Each model runs each scenario $10$ times to account for response variability, $9 \times 10 \times 4 = 360$ runs in all. A failed LLM call is retried up to three times with the rejection message appended. Runs that still fail count as execution errors. Accuracy metrics are computed on completed runs and scored deterministically, by set equality on identifiers and exact match on the cause type, with no paraphrase matching and no LLM-as-judge.
\begin{itemize}
 \item \textbf{Root cause accuracy} counts the runs whose \texttt{rootCauseIds} equals $R^*$ as a set, which penalises a missed co-cause as much as a named healthy resource.
 \item \textbf{Evidence exact match} counts the runs whose \texttt{evidenceIds} equals $E^*$ exactly.
 \item \textbf{Cause-type accuracy} counts the runs whose \texttt{rootCauseType} equals $T^*$.
 \item \textbf{Full diagnostic accuracy} counts the runs where all three are correct at once.
\end{itemize}
A static analyser also checks three invariants on every completed trace.
\begin{itemize}
 \item \textbf{Hallucination rate} is the fraction of identifiers mentioned by the agent that no prior tool call returned.
 \item \textbf{Protocol compliance} is the fraction of runs whose tool calls follow the five steps in order.
 \item \textbf{Tool misuse} is the fraction of tool calls outside Definition~\ref{def:mcp-interface} or violating its signatures.
\end{itemize}

\subsection{Results}
\label{sec:results}

\noindent Table~\ref{tab:bench-results} gives diagnostic accuracy by model and category, Table~\ref{tab:faithfulness-results} faithfulness and Table~\ref{tab:cost} inference cost.

\begin{table}[ht]
\centering
\scriptsize
\setlength{\tabcolsep}{3pt}
\resizebox{\columnwidth}{!}{%
\begin{tabular}{lcccc}
\toprule
\textbf{Model} & \textbf{Root cause} & \textbf{Evidence match} & \textbf{Cause type} & \textbf{Full diag.} \\
\midrule
Claude Opus 4.7  & 100/100/100/\textbf{100} & 100/90/100/\textbf{97}    & 100/100/100/\textbf{100}     & 100/90/100/\textbf{97} \\
Claude Haiku 4.5 & 100/63/100/\textbf{88}   & 100/90/100/\textbf{97}    & 100/100/97/\textbf{99}       & 100/63/97/\textbf{87} \\
Llama 3.3 70B    & 97/34/97/\textbf{76}     & 97/48/97/\textbf{81}      & 100/97/100/\textbf{99}       & 97/10/97/\textbf{68} \\
GPT-OSS-120B     & 70/47/73/\textbf{63}     & 70/43/70/\textbf{61}      & 70/80/70/\textbf{73}         & 70/27/70/\textbf{56} \\
\bottomrule
\end{tabular}}
\caption{Diagnostic accuracy by model, in percent rounded to the nearest integer. Each cell gives A/B/C/\textbf{All} over the completed runs, $30$ per category and $90$ in all, except for Llama 3.3 70B with one failed run in B and one in C ($29$, $29$ and $88$).}
\label{tab:bench-results}
\end{table}

\begin{table}[ht]
\centering
\small
\resizebox{\columnwidth}{!}{%
\begin{tabular}{lccc}
\toprule
\textbf{Model} & \textbf{Hallucination rate} & \textbf{Protocol compliance} & \textbf{Tool misuse} \\
\midrule
Claude Opus 4.7   &  0.00\% & 100.0\% &  0.00\% \\
Claude Haiku 4.5  &  0.00\% & 100.0\% &  0.00\% \\
Llama 3.3 70B     &  0.00\% & 100.0\% &  0.00\% \\
GPT-OSS-120B      &  2.11\% & 100.0\% &  0.00\% \\
\bottomrule
\end{tabular}}
\caption{Faithfulness by model, over the completed runs of each model ($90$, or $88$ for Llama 3.3 70B).}
\label{tab:faithfulness-results}
\end{table}

\begin{table}[ht]
\centering
\small
\resizebox{\columnwidth}{!}{%
\begin{tabular}{lrrrrr}
\toprule
\textbf{Model} & \textbf{Tok.\ in} & \textbf{Tok.\ out} & \textbf{Total} & \textbf{Iter.} & \textbf{Full diag.} \\
\midrule
Claude Opus 4.7  & 20\,555 & 1\,612 & 22\,166 & 5.0  & 96.7\% \\
Claude Haiku 4.5 & 15\,833 & 1\,346 & 17\,179 & 5.0  & 86.7\% \\
Llama 3.3 70B    & 14\,662 &   441  & 15\,103 & 6.2  & 68.2\% \\
GPT-OSS-120B     & 22\,069 & 1\,086 & 23\,155 & 11.1 & 55.6\% \\
\bottomrule
\end{tabular}}
\caption{Inference cost per run, averaged over the completed runs of each model ($90$, or $88$ for Llama 3.3 70B). Tokens are those reported by the provider API over all turns of a run. \textbf{Iter.} is the mean number of agent turns.}
\label{tab:cost}
\end{table}

Two of the four models diagnose nearly every run correctly. Claude Opus 4.7 names the root cause resources and the cause type on all $90$ runs. Its full diagnostic accuracy of $96.7\%$ falls short of $100\%$ only on three runs of B2, where it also cites the healthy rack notes of \texttt{AppNode-B} and \texttt{AppNode-C} as eliminative evidence. The exact match on evidence counts these as errors. Claude Haiku 4.5 reaches $86.7\%$. The agent stays grounded with every model, at $100\%$ protocol compliance, $0\%$ tool misuse and at most $2.1\%$ hallucination on mentioned identifiers (Table~\ref{tab:faithfulness-results}). Residual error rates are $0\%$ except for Llama 3.3 70B at $2.2\%$. A run consumes $15$k to $23$k tokens (Table~\ref{tab:cost}). More tokens do not buy accuracy. GPT-OSS-120B consumes the most and is the least accurate, its $11.1$ turns against $5.0$ for the Anthropic models being re-reads, not additional evidence.

The models differ once the evidence is spread across resources. On categories A and C, where the discriminating fragment sits at a single node, all four score between $70\%$ and $100\%$ full diagnostic accuracy. Category B requires combining fragments across resources. Every model drops there, Opus 4.7 to $90\%$, Haiku 4.5 to $63\%$, GPT-OSS-120B to $27\%$ and Llama 3.3 70B to $10\%$. The ranking shifts with it. Llama 3.3 70B, ahead of GPT-OSS-120B overall at $68.2\%$ against $55.6\%$, falls behind it on B. Since the protocol reads every resource, this drop measures how well a model combines clues gathered across resources. It says nothing yet about the choice of what to read.


\section{Discussion}
\label{sec:discussion}

This paper shows that structure-only methods cannot address the regime of Section~\ref{sec:decomposition} and that ReAct, the most common agentic architecture, is a feasible way to interpret the data it requires. It does not show that an agentic layer is necessary, nor that ReAct is the best choice. On our graph the protocol reads every resource of $\sigma(s)$. A single model call over all the collected fragments might do as well there. The case for an agentic layer rests on larger infrastructures, where the agent would have to choose which resources to read at all as its hypotheses form. Other agentic architectures divide the work between the model and the code around it differently. Whether such a layer is needed and which architecture suits diagnosis over an infrastructure graph as it grows are the questions this work leaves open.

A method that could contest our results has to read the same unstructured data. The simplest collects every fragment of $\sigma(s)$ and asks a single model call for the diagnosis. Retrieval-augmented pipelines select the fragments in one shot by similarity to the incident and reason over the batch. That suits cases where the relevant fragments can be identified from the query alone, which regime (i) of Section~\ref{sec:decomposition} makes unlikely. A neuro-symbolic pipeline externalises that data into features before a classical algorithm runs. It meets the same difficulty one stage earlier, in deciding what to externalise. We implemented none of them and our results do not measure them. Comparing them with an agentic layer on graphs too large to read in full is the natural next experiment.

The main limitation is the benchmark itself, synthetic and small, with nine scenarios, one query template each and no adversarial fragments. Ten runs per scenario measure how much a model varies, not how far the results generalise, which only real incidents can show. The protocol treats an event describing an already-resolved problem like an active one. A large $\sigma(s)$ may return more fragments than the context window holds. Server-side prefiltering would displace that problem without removing the need to interpret. It is also confined to one telecom and datacenter ontology. Other domains whose evidence sits in unstructured data attached to a graph, such as industrial maintenance, remain untested.

Letting operators write a note instead of driving a complex data model makes the diagnosis depend on how a model reads that note. Heavily structured pipelines are not error-free either, since rich data models accumulate modelling mistakes and the shortcuts operators take when a comment is faster than the schema. An operational deployment would still need a per-run confidence signal that decides whether an output can be acted on or should go to a human, a design we leave to follow-on work.


\section{Conclusion}

Root cause analysis over an infrastructure digital twin often turns on unstructured data attached to the graph, whose relevance only emerges during the investigation. No method restricted to the structured attributes can be right on every such instance, since two instances may differ only in that data. An agentic layer can interpret it. With ReAct as its instance, a frontier-tier model names the root cause and its type on every run of our benchmark and reaches $96.7\%$ end-to-end diagnostic accuracy, a mid-tier model $86.7\%$, with no tool misuse and almost no hallucinated identifier. Accuracy drops for every model when the evidence must be combined across several resources. In this regime, reading that data is what makes the diagnosis possible. An agent is one feasible way to read it.

\section*{Code Availability}
The implementation of the agent and its MCP server and the nine benchmark scenarios will be released upon publication.

\bibliographystyle{IEEEtran}
\bibliography{references}

\end{document}